Я. А. ПЕРВАНОВ

# ЗАМЕТКИ ПО ЭЛЕКТРОННОЙ ЛЕКСИКОГРАФИИ

> *Мы полагаем, что электронный словарь - это особый лексикографический объект, в котором могут быть реализованы и введены в обращение многие продуктивные идеи, не востребованные по разным причинам в бумажных словарях [В. Селегей].*

Эти заметки являются продолжением темы, затронутой в статье В. Селегея *Электронные словари и компьютерная лексикография.* Речь пойдет о том, каким может быть электронный словарь, имеющий в качестве объекта описания близкородственные языки. Очевидно, такая постановка вопроса допускает несколько вариантов ответов, а единственным подтверждением правильности каждого из них мог бы стать словарь или словари - действующие продукты, созданные в электронном виде, существующие на электронном носителе, предназначенные для электронного чтения и имеющие перспективы электронного развития. Мы постараемся обойтись без противопоставления "традиционный – новый", "классический – современный" и т.д., потому что об этом написано немало и не только по поводу бумажных и электронных словарей.

Все знают, чем является в обыденном понимании электронный словарь. У него, в отличие от книги, нет предметного образа, его нельзя "подержать в руках", его страницы не пронумерованы, его можно включить и выключить, он сделан в виде программы, его читают на дисплее, его необходимо инсталлировать и "листать" мышью, одним пальцем или электронной "ручкой" и т.д.

Перечисленные атрибуты электронного словаря на самом деле являются мнимыми. Виной тому – красивое название *электронный словарь*, в котором прилагательное, обозначающее первоначально узкую техническую предметно-понятийную сферу, приобрело резонансную частоту употребления и не "по рангу" определяет словарь как особый вид *текста*.

> *Термин "электронный словарь" стал уже привычным. При этом атрибут "электронный" характеризует свой объект настолько же поверхностно, насколько противоположный ему атрибут "бумажный" - традиционные словари [В. Селегей: электронный ресурс].*

Одна из причин состоит в противопоставление понятий *бумажный – электронный,* которое, набрав силу несколько десятилетий назад, продолжает упорно затенять тот факт, что текст (образ) и носитель являются несвязанными сущностями.

## Языковой резонанс

[Резонанс](#) — явление, заключающееся в том, что при некоторой частоте вынуждающей силы колебательная система оказывается особенно отзывчивой на действие этой силы. Лексическая синонимия – вид семантического резонанса. "Раскачивается" понятие, ищущее выражения в языке[1]. Вынуждающая сила – желание выразить мысль точнее, лучше, богаче. В отличие от механического резонанса, семантический резонанс высвобождает энергию вторичного означивания "благодаря механизму установления связей между прецедентным феноменом и новым контекстом его реализации" [Бойко 2011: электронный ресурс].

Итак, *электронный словарь* – это электронным образом представленный "книжный" словарь или совокупность таковых, подобно тому как *электронный документ* является электронной версией "бумажного" документа. Так ли это? Нетрудно догадаться, что прецедентным феноменом в данном случае является не первичная "книжная" версия, несмотря на частые попытки подтолкнуть читателя именно к этой мысли. Прецедентным феноменом являются понятия *словарь* и *документ,* проще говоря – некий *новый структурированный текст, имеющий определенный объем, цель и несущий определенную идею.* Понятие и образ можно воссоздать на бумаге или на экране компьютера, на песке или белом полотне, его можно вышить или выжечь – смена носителей в данном случае не играет решающей роли. Что останется от электронного словаря, если в ближайшем будущем удастся создать *бумажный биоэлектронный словарь*?

Мы согласны с В. Селегеем, что электронный словарь – это особый лексикографический объект. Должно соблюдаться, на наш взгляд, одно условие: этот объект необходимо создавать, а не просто воспроизводить. Качественное отличие электронного словаря от бумажного видится не столько в электронной разметке и гипертексте, в карточках и окнах на месте бумажных страниц и словарных статей, сколько в новой манере лексикографической интерпретации языковой семантики. Эта интерпретация вовлекает читателя в виртуальную лексикографию совершенно иным образом.

> ***Задача создания такого словарного содержания, которое позволило бы сделать единицей анализа отдельное лексическое значение, а не морфологическую лексему, видится нам наиболее перспективным направлением в компьютерной лексикографии [В. Селегей].***

По-видимому, В. Селегей имеет в виду лексическое значение в форме толкования или определения (пусть формализованного в рамках той или иной модели), поскольку, как известно, массовому читателю можно предложить лишь что-то, напоминающее "наивное" толкование (резонанс интерпретаций) которое он в состоянии дать сам в рамках своей лингвистической компетенции.

Нам трудно представить себе , как можно сделать электронный словарь более "лингвистическим", нежели бумажный, если в работе над электронным массовым словарем будет делаться упор лишь на электронные средства (словари типа Wordnet являются исключением): наращивание инструментальной части - базы данных, увеличение скорости получения ответа по разным критериям поиска, новая маркировка, аудио и т.д. Очевидно, электронный словарь должен создавать новый интерактивный резонанс восприятия словарной информации. Он должен мультиплицировать эффект семантической интерпретации и вовлечь читателя в новую лингвистическую интригу. Это так же сложно, как и создание бумажного словаря нового концептуального типа.

Возьмем следующий тривиальный пример: *Яблоко висит на ветке яблони.* В терминологии некоторых исследователей это будет псевдопредложением [Звегинцев, 1976, с.185-186]. С точки зрения коммуникативного назначения, или интенции, это предложение относится к прагматическому типу констативных, содержащих "нечто вроде [я утверждаю]" [Почепцов, 1975, с.181]. Однако любой взрослый носитель языка отметил бы также его тавтологичность. Человек, листающий толковый словарь русского языка, удивился бы, если бы узнал, что предложение *Яблоко висит на ветке яблони* [яблоко, висеть, ветка, яблоня] можно "истолковать": ['плод', который 'находится в вертикальном положении, на весу, без опоры' на 'небольшом боковом отростке, побеге' `фруктового дерева из семейства розовых'] (СО). Однако он вряд ли удивится, если мы объясним ему, что здесь не одно, а как минимум четыре утверждения: 1. Яблоко на ветке яблони может только *висеть* (не *возвышаться*, *стоять* и т.д.); 2. На ветке яблони может висеть только то, что называется *яблоко* (не *вишня); 3. Яблоко может висеть только *на ветке яблони* (не *на стволе* или *под веткой*, хотя реально оно находится под веткой). 4. Яблоко может висеть только на ветке *яблони* (не *вишни).

"Сказано то, что сказано" - так можно определить эту замкнутую логическую цепь предложения, отражающую лингвистический аспект наших знаний о ситуации. Тот факт, что исходное предложение воспринимается как тривиальное суждение и может вызвать у собеседника ряд нежелательных реакций, является свидетельством того, что мы вторгаемся в сферу "языковых примитивов" и игнорируем резонансный характер языкового значения. Продолжая преодолевать сопротивление языка и все дальше тавтологизируя его, мы употребили бы (в качестве отрицательного материала) выражения *\*яблоко яблони*, *\*ветка ствола*, *\*яблоко ветки* или *\*рука тела,* *\*нос лица* и т.д. Запрет на такие выражения можно объяснить закономерностями сочетаемости, соположенностью предметов, понятием части и целого и т.д. Эти объяснения можно дополнить тем соображением, что язык имеет своего рода защитный механизм, который срабатывает по мере приближения к элементарным схемам, хранящимся в "невыразимом" агрегатном состоянии. Вместо вышеуказанных словосочетаний мы чаще прибегаем к иносказаниям: *плод яблони, рука человека* или *конечность тела*, *пальцы руки* (не * *ладони*), *нос человека* или *нос - на лице* (не \**нос лица, нос головы*) , *ветка яблони, яблоко на ветке, яблоко с ветки* и т.д.

Представляется, что более удачным толкованием слова *яблоко* было бы не 'плод яблони', а 'плод дерева, которое называется *яблоня*', но, приводя пример с яблоком, мы преследовали иную цель. Прецедентный феномен в собственном концептуальном контексте расценивается говорящим как схематическое deja-vu, иными словами, лишен резонанса. Электронный словарь, если он претендует на роль семантического каталога нового типа, должен заставить элементарные единицы анализа резонировать по-другому. Лучший способ (он же и неформальный) – вовлечь читателя в механизм интерпретации языковых значений.

**Модель двуязычного интерактивного словаря**

Двуязычный словарь является реляционным: его объект не тождествен одному языку, средством описания языка А выступает другой произвольно взятый язык В, который может быть в свою очередь описан языком А (однонаправленные переводные словари). Объектом микроструктуры словаря являются номинально лексемы, однако по сути выстраивается бинарное отношение выражения смыслов. Это отношение рефлексивно, оно устанавливает подобие и как правило не может быть сведено к эквивалентности типа слово – слово:

*Фиг 1. Схематическое представление однонаправленного двуязычного словаря*

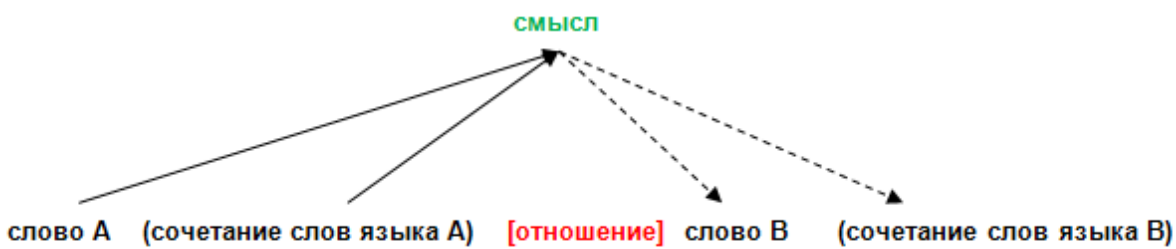

При таком раскладе стрелки, ведущие к В, могут вести также в никуда: типичным случаем являются лакуны, или, по крайней мере, их трактовка в теории перевода и лингвокультурологии.

Создание двуязычного словаря в компьютерной среде может следовать канонам двуязычной лексикографии: в этом случае уместно говорить об электронной *версии* словаря. Как правило, речь идет о переводном словаре со всеми присущими ему достоинствами и недостатками. Мы утверждаем, что компьютерные технологии позволяют строить двуязычный словарь на разных принципах, одним из которых является **интегральный сопоставительный** принцип. Первые робкие попытки назвать двуязычный словарь сопоставительным – ср. термины *дифференциальный словарь, контрастивная лексикография, сравнительно-сопоставительный словарь, толково-сопоставительный словарь* и т.д. – и тем самым провести разграничение в методах (мы надеемся, что речь не идет о терминологической моде) уже являются фактом, но говорить о сопоставительной лексикографии как особом жанре, по-видимому, еще рано.

К настоящему времени одна лишь идея создания сопоставительного (не просто переводного) словаря могла бы обескуражить многих. Если в переводном словаре по традиции задано направление: от неродного к родному или наоборот, то в сопоставительном словаре понятие «родного» терминологически бесполезно, а направление как таковое – теряет свою «переводческую» подоплеку. Если, например, в двуязычном словаре болг. *заблуждавам* имеет эквиваленты *обманывать, вводить в заблуждение*, и на этом сопоставление заканчивается, то в сопоставительном словаре – это лишь начало, поскольку необходимо определить, во-первых, симметричный или асимметричный характер соотношения эквивалентов на фоне других эквивалентов типа *обманывать – лъжа, лъжа – лгать, вводить в заблуждение – вкарвам в заблуда* и т.д., во-вторых, представить все эти эквиваленты в упорядоченной классификации, в третьих, найти новый способ лемматизации материала (пар) в словарной статье. Это лишь некоторые различия. В сопоставительном словаре, в отличие от переводного, необходимо отражать толкования значений многозначных слов, а не полагаться на интуицию читателя или лингвиста. Далее, необходимо включить обширные цитаты из национальных корпусов, было бы неплохо учесть синонимы, фразеологизмы и перифразы.

Ситуация может оказаться намного сложнее, если задаться целью не просто сопоставлять по алфавитному порядку произвольные группы слов, а включить идею антропоцентризма в сопоставительное описание лексики двух языков. В таком случае словарь неизбежно приобретет элементы идеографического или семантического словаря, но придется также подумать об объеме словаря, о подготовленности читателя, а главное – коренным образом изменить макроструктуру словаря и сделать его интерактивным.

Очевидно, создание подобного словаря сопряжено с необходимостью ответить на ряд вопросов, среди которых центральное место занимают вопросы о структуре, методе описания и предназначении словаря. Несомненно, такой словарь невозможно создать в виде большого проекта, без предварительной проработки экспериментального словаря небольшого размера, проверки его функциональных возможностей, технических решений, опроса читателей и маркетинга.

Идея составителей нового электронного [русско-болгарского словаря СЭД](#) проста [см.: Перванов, 2010]. Читателю необходимо предложить интерактивный словарь нового образца, в котором эквиваленты представлены как развернутые цепи лексических пар, слова сами поочередно "переводят" друг друга, языки как бы взаимно "отражаются", а на фоне переводных эквивалентов "видны" их синхронизированные толкования, синонимы, фразеологизмы. В известной мере такая словарная статья должна показывать живую динамику межъязыковых соответствий, она разворачивается "стереоскопически" сверху вниз и от центра к периферии, а основными элементами являются узловые пары, скрепленные символом эквивалентного отношения. Несколько словарей в одном и на одной странице - это замечательно. Подобный словарь мог бы стать ценным подспорьем переводчику, преподавателю языка, студенту, но также лингвисту, благодаря полному каталогу типов семантических соответствий слов в двух языках.

Сопоставительный интерактивный словарь СЭД является необычным лексикографическим проектом. В силу ряда нестандартных решений его структура и содержание могут поначалу повергнуть в уныние неподготовленного читателя. С другой стороны, словарь ждет проверка не только со стороны доброжелательных критиков, но и тех читателей, которых В. Селегей называет "особо вредными пользователями", чье "маниакальное стремление обнаружить ошибку или лакуну" создает немало хлопот авторам. В словаре СЭД нет догмы, он открыт для критики и активного участия всех, кто интересуется тем, *что* стоит за словом и его употреблением в речи.

Цель настоящего словаря – предложить один из путей приближения к реальному прототипу сопоставительного электронного русско-болгарского словаря нового образца, в котором накопленные знания о сценариях говорящего и наивной «картине мира» (схемы, фреймы, гештальты) были бы надлежащим образом синхронизированы с лексикографической интерпретацией значений и научно обоснованным сопоставлением двух или более языков.

## Читатель – каким он может быть?

По справедливому замечанию Р. Хартмана, между составителями словарей и целевой читательской аудиторией как правило нет прямого контакта [Hartmann, 2001, с. 24]. Обобщая роли участников процесса подготовки и использования словаря, Р. Хартман отмечает, что теоретически составитель словаря, критик-исследователь, преподаватель и читатель должны поддерживать как минимум попарную двунаправленную связь, но на практике существует немало коммуникативных препятствий, которые делают эту связь трудно осуществимой [там же, с. 26].

Другая проблема связана с категоризацией семантики в двуязычном словаре и реальными потребностями переводчика, поскольку "переводы осуществляются на уровне текстов, а не на уровне словарей, да еще и столь ограниченных по объему и по своим возможностям, как словари двуязычные" [Павлова, 2011: электронный ресурс]. Реплика:

> ***It is worth reminding the reader that translation is creative work, 'variation on the theme', whereas dictionaries are based on consistent confrontation of lexical systems [L. Minaeva, 2007, c. 93].***

По-видимому, невозможно создать "словарь золотой середины". Это звучит еще более актуально для читателей сопоставительного электронного словаря, которые не видели ничего, кроме переводного (хорошего, разумеется), строящего словарную статью по линейному принципу **А переводится как ... либо как ..., а иногда и как ...** Значит ли это, что читатель обречен не понимать иного способа представления межъязыковых эквивалентов?

Отнюдь нет. Читатель имеет право знать, *что* ему предлагают, и сделать свой выбор в той или иной проблемной или учебной ситуации. По правилу, словарная страница интерактивного сопоставительного словаря должна быть в меру избыточной. Это значит, что объем информации, которую читатель **может** получить при посещении страницы, зависит прежде всего от объема информации, которую он **хотел бы** получить при определенных технических возможностях словаря. Кроме толкований, синонимов, фразеологизмов и иллюстративных примеров, на страницах словаря СЭД представлена сопоставительная классификация эквивалентных связей слов в двух языках. Сверхзадачей словаря является когнитивный анализ макрогруппы слов в двух языках. Как следует ожидать, не все модули словаря будут одинаково интересны и специалистам, и неискушенным читателям.

Разумеется, плох тот словарь, который не объясняет, как им пользоваться. Здесь тоже не существует правила "золотой середины". Оперативной базой словаря СЭД является сайт http://www.sedword.com/. Большая часть ресурсов сайта посвящена различным аспектам словаря. Мы надеемся, что читатель оценит это по достоинству.

## Заключение

Работа над интерактивным русско-болгарским словарем продолжается. Окончательный вариант словаря может быть существенно изменен. Главный вывод заключается в том, что электронная лексикография, несомненно, имеет свой объект, цели, задачи и перспективы. Необходимо сближать лингвистические и электронные составляющие словарного дела и подключить массового читателя к словарю на этапе разработки. Тогда будут и новые мехи, и новое вино. Т.е. лексикографический резонанс.

\* \* \*

*Ключевые слова:* электронный словарь, "бумажный" словарь, языковой резонанс, компьютерная лексикография, модель двуязычного словаря, сопоставительный словарь.

# ЛИТЕРАТУРА

---

[1]) С.Д. Кацнельсон характеризует понятийное поле как «противопоставление понятии, ищущее выражения в языке» [Кацнельсон,1965, с. 77].выражения в языке» [Кацнельсон,1965, с. 77].